\documentclass{article}
\usepackage[parfill]{parskip}    
\usepackage{graphicx}
\usepackage{amssymb}
\usepackage{amsmath}
\usepackage{spconf}                      
\usepackage{subfigure}

\title{Convex Approaches to Model Wavelet Sparsity Patterns}
\twoauthors{Nikhil S. Rao, Robert D. Nowak, Stephen J. Wright}{University of Wisconsin-Madison}{Nick G. Kingsbury}{University of Cambridge}

\begin{document}
\ninept
\maketitle

\begin{abstract}
Statistical dependencies among wavelet coefficients are commonly represented by graphical models such as hidden Markov trees (HMTs).  However, in linear inverse problems such as deconvolution, tomography, and compressed sensing, the presence of a sensing or observation matrix produces a linear mixing of the simple Markovian dependency structure.  This leads to reconstruction problems that are non-convex optimizations. Past work has dealt with this issue by resorting to greedy or suboptimal iterative reconstruction methods.  In this paper, we propose new modeling approaches based on group-sparsity penalties that leads to convex optimizations that can be solved exactly and efficiently. We show that the methods we develop perform significantly better in deconvolution and compressed sensing applications, while being as computationally efficient as standard coefficient-wise approaches such as lasso.
\end{abstract}

\keywords
wavelet modeling, deconvolution, compressed sensing

\section{Introduction}
\label{intro}
Statistical dependencies among wavelet coefficients are commonly represented by trees or graphical models such as hidden Markov trees (HMTs) \cite{crouse98}. HMTs provide superior denoising results compared to independent coefficient-wise thresholding/shrinkage methods, like the lasso \cite{tibshirani}.   Fast exact and/or approximate inference algorithms exist in many situations, but not all.  In linear inverse problems (e.g., deconvolution, tomography, and compressed sensing) the presence of a sensing/observation matrix can linearly mix the Markovian dependency structure so that simple and exact inference algorithms no longer exist.  Past work has dealt with this issue by resorting to greedy or suboptimal iterative reconstruction methods such as those based on belief propagation \cite{schniter10}, iterative re-weighting \cite{duarte08}, or variants of the Orthogonal Matching Pursuit \cite{huang09,lado} (see also \cite{modelbased, model09}). In this paper, we propose a new modeling approach based on group-sparsity penalties that leads to convex optimizations that can be solved exactly and efficiently.  Our results show that the approach performs much better in deblurring and compressed sensing applications, while being as computationally efficient as standard coefficient-wise approaches. Our work uses the group lasso with overlap formulation introduced in \cite{jacob}, which we further modify to better represent dependencies among wavelet coefficients. Note here that we use the $\ell1-\ell2$ norm formulation. Similar work could be performed by using the $\ell1-\ell\infty$ norm \cite{l1linf}.

We motivate our problem in section \ref{pform}. In section \ref{model}, we explain how we model the wavelet transform coefficients into overlapping groups. Section \ref{results} outlines the experiments we performed, and the results obtained. We conclude the paper in section \ref{conc}.

\section{Problem Formulation}
\label{pform}
Consider a linear observation model which can represent blurring, tomographic projection, or compressed sensing:
\begin{eqnarray*}
y = L \, x \ + \ w \ ,
\end{eqnarray*}
where $y$ is the measured data, $L$ is a linear observation operator, $x$ is the image to be reconstructed, and $w$ is additive Gaussian noise. Throughout the paper we will assume a standard matrix-vector representation: the image is represented as a column vector and linear operators are represented as matrices.  Images typically have approximately sparse representations in the wavelet domain, and this has led to many approaches to image reconstruction that attempt to exploit this property \cite{mario03,sparsa}.  For example, the standard $\ell_2/\ell_1$ or {\em lasso} \cite{tibshirani} reconstruction problem is written as
\begin{eqnarray}
\label{lasso}
\widehat{\theta} & := & \arg \min_\theta \left\{ \, \frac12 \|y-A\theta\|_2^2 \, + \, \lambda_l \|\theta\|_1 \, \right\} \,  
\end{eqnarray}
where $A:=LW$, the composition of $L$ and the inverse wavelet transform $W$, $\theta$ denotes a set of wavelet coefficients, and $\lambda_l>0$ is a regularization parameter that balances the tradeoff between fitting to the data $y$ and minimizing the $\ell_1$ norm of $\theta$ (which serves as a surrogate for sparsity).

The lasso penalty reflects the fact that the wavelet coefficients are approximately sparse, but in reality not all patterns of sparsity are equally plausible/probable. A commonly observed effect is the persistence of large (or small) wavelet coefficient across scales due to the localized nature of edges.  Many models have been proposed to represent such patterns, and in particular tree-structured models have been among the most successful and widely used (e.g., hidden Markov trees\cite{crouse98}).  Tree-structured models admit very efficient estimation procedures based on pruning or message-passing algorithms in denoising applications where $A$ is the identity operator, but when $A$ is not identity such simple strategies can no longer be applied. In fact, in general the optimization problem resulting from tree models is non-convex (unlike equation \eqref{lasso} above) and so exact solutions are difficult or  almost impossible to obtain. This issue has been addressed by resorting to greedy or suboptimal iterative reconstruction procedures \cite{schniter10,duarte08,modelbased,model09,huang09,lado}.

This motivates the main idea and contribution of this paper.  While tree models can represent the patterns of sparsity (or approximate sparsity) in the wavelet coefficients of natural images, they are not the only way to capture such effects.  We are particularly interested in modeling the so-called ``parent-child dependency'' \cite{crouse98}, which is used in reference to the persistence of large/small wavelet coefficients across scales.
Specifically, if a wavelet coefficient at a certain spatial location and scale is large/small, then its ``neighboring'' coefficients at roughly the same location but finer or coarser scales tend to be large/small. The term parent-child refers to a pair of coefficients at a certain location and adjacent scales. Our goal is to exploit the fact that the coefficients in each pair are typically both large or both small (or zero) in magnitude.  This can be accomplished using an overlapping-group penalty function \cite{jacob,jenatton,jenatton10} that generalizes the lasso in a way that captures parent-child dependencies while at the same time retaining its convex nature.  

\begin{figure}[htp]
\begin{center}
\subfigure[]{
\includegraphics[scale =0.5]{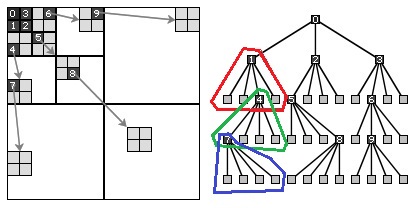}
\label{PC4}} 
\subfigure[]{
\includegraphics[scale =0.5]{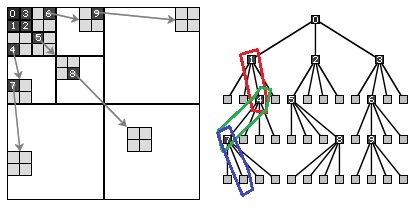}
\label{PC2}} 
\caption{Quadtree corresponding to the 2-d DWT. At each scale, parent coefficients can be grouped with child coefficients.  All four children may be grouped together with the parent (above) or the parent can be grouped with each child individually (below).}
\label{wtree2}
\end{center}
\end{figure}

Fig.~\ref{wtree2} depicts wavelet quadtree structures and example of parent-child groups.  Each (non-leaf) coefficient has an associated orientation (horizontal, vertical or diagonal) and four child coefficients of the same orientation at the finer scale below it. 
Many options exist for grouping parents with children, and two options are depicted in the figure.  Many other grouping schemes are also possible in our framework (e.g.,  including ``grandparent" coefficients as well), but we will not explore such extensions in this paper.

The main contributions of the paper are threefold: 1) We introduce a new approach to representing the wavelet coefficient sparsity patterns commonly observed in natural images; 2) We adapt and extend recently proposed methods for group lasso with overlaps in order to take advantage of these sparsity patterns in linear inverse problems using simple convex optimization techniques; 3) We demonstrate fast and efficient reconstruction in image deblurring and compressed sensing and significant improvements in reconstruction error relative to the lasso.

\section{Convex Group Regularizers}
\label{model}

To encourage solutions that have wavelet coefficient sparsity patterns reflective of the parent-child group structure and persistence of large/small coefficients across scales we apply the group lasso regularization \cite{yuanlin}:
\begin{align}
\label{glasso}
\widehat{\theta}_{gl} &:=  \underset{\theta}{\operatorname{argmin}} \left\{ \, \frac{1}{2}\|y-A\theta\|_2^2 \, + \, \lambda_g \sum_{g \in \mathcal{G}} \|\theta_g\|_2 \, \right\} \ , 
\end{align}
where $\mathcal{G}$ denotes the collection of all parent-child coefficient groups and
$g$ denotes one such group. The quantity $\theta_g$ is the subvector of $\theta$ of the coefficients in group  $g$. The group lasso penalty enforces group sparsity by setting whole groups to be zero if the $\ell_2$ norm of the group is small relative to their importance in the data-fitting term. In most applications of the group lasso, the groups are assumed to be disjoint, but in our case they are overlapping and hence the penalty terms are coupled.  Because of this coupling the standard group lasso optimization strategies (e.g., \cite{sparsa}) cannot be directly applied.
We offer two approaches to deal with this issue.

\vspace{-2 mm}
\subsection{Variable Replication Approach}
One way to deal with overlapping groups is to introduce replicates of each coefficient so each group involving a certain coefficient has its own ``copy'' of it.  This ``decouples" the overlapping groups from each other. This approach was proposed and analyzed in \cite{jacob}. The replication of variables (and subsequently the columns of the matrix A) results in  a formulation that can be expressed as
\begin{align}
\label{oglr}
\widehat{\theta}_{\tiny OGLR} :=  \underset{\theta}{\operatorname{argmin}} \left\{ \, \frac{1}{2}\|y - \tilde{A}\tilde{\theta}\|_2^2 
+ \lambda_{rep} \sum_{\tilde{g} \in \tilde{\mathcal{G}}} ||\tilde{\theta}_{\tilde{g}}||_2 \right\} \ ,
\end{align}
where $\widetilde \theta$ is the extended vector with replicates and $\tilde{A}$ is a matrix obtained by replicating the corresponding columns of $A$.  Because the penalty function is now separable,  computationally efficient iterative shrinkage/thresholding methods can be applied \cite{sparsa}. 

\vspace{-2 mm}
\subsection{Constraint-Based Approach}
The overlap group lasso with the replication strategy treats each group independently of each other. This means that, we can have a grandparent - parent coefficient group selected, but not the parent - child group, violating the persistence of wavelet transforms across multiple scales \cite{crouse98}. This motivates the use of a penalty that tends to cause all the coefficients in a location across scales to have a similar value. To this end, we modify equation \eqref{oglr} as
\begin{eqnarray}
\notag
\widehat{\theta}_{OGL}&:=  \underset{\theta}{\operatorname{argmin}} \Bigg\{ \, \frac{1}{2}\|y - A\theta\|_2^2 
+ \lambda_{ogl} \sum_{\tilde{g} \in \tilde{\mathcal{G}}} ||\tilde{\theta}_{\tilde{g}}||_2 .  \\
\label{moglasso}
&  \ \hspace{.7in} +  \ \frac{1}{2} \tau^2 \, \sum_{i=1}^n \sum_{j \in J_i} (\theta_i - \theta^{(j)}_i)^2 \,  \Bigg\} 
\end{eqnarray}
where $\theta_i$ is the ``master copy" of the $i$-th coefficient and $\theta_i^{(j)}$ denote the copies of it that appear in the group penalty terms. $J_i$ is the set of replicated variables of $\theta_i$.  Setting $\tau>0$ large forces the replicated copies to agree, yielding a solution to the group lasso in  (\ref{glasso}). This encourages a stronger degree of persistence across scales. To the best of our knowledge, this approach has not been previously proposed for the overlapping group lasso problem. Note that the additional quadratic penalty can be combined with the quadratic data-fitting term to obtain a quadratic plus separable group sparsity penalty, and we can then directly apply standard solvers such as \cite{sparsa}. Henceforth we use L to denote lasso, OGL to denote the Overlap Group Lasso (corresponding to \eqref{glasso} and equation \eqref{moglasso} with $\tau\gg 1$) and OGLR to denote the overlap group lasso with replication (equation \eqref{oglr}).

\vspace{-2 mm}
\subsection{Sparsity Patterns and Penalties}
\begin{figure}[htp]
\begin{center}
\subfigure[Haar DWT of the Cameraman]
{\includegraphics[scale = 0.2]{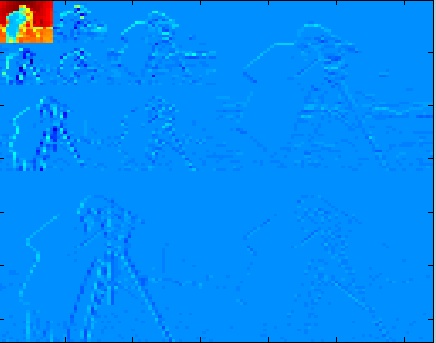}
\label{unjumbled}}
\subfigure[Scrambled DWT]
{\includegraphics[scale = 0.2]{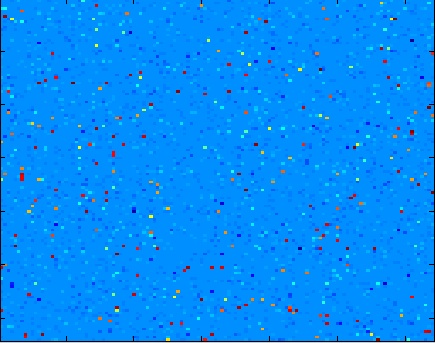}
\label{jumbled}}
\caption{Ratio of sparsity penalty of (a) to (b): lasso 1.00; group lasso 0.70; group lasso with replication 0.85. The group penalties significantly favor the structured sparsity pattern of (a)} 
\label{normComp}
\end{center}
\end{figure}

To demonstrate the effect of the group penalties in contrast to the usual $\ell_1$ penalty, we compare their values for the `cameraman' image.  
Fig. \ref{normComp} shows the wavelet coefficients for the image in the standard organization (left) and in a randomized organization (right).  The $\ell_1$ norm is invariant to the randomization, but the group penalties do change because parent-child dependencies are not preserved.  To quantify the degree to which the group penalties encourage parent-child persistence, we compute the ratio of the group penalties for the left image to the randomly organized image on the right.  The group penalties are larger when the parent-child relationships are lost due to randomization, and so the ratios are less than 1 for the group penalties.  The OGL penalty ratio is smaller than the OGLR (with replication) penalty, indicating that the OGL penalty  in (\ref{glasso}) more strongly favors the structure in this image compared to the OGLR penalty (\ref{oglr}) or the $\ell_1$ penalty in the lasso.  Next we show  experimental evidence that OGL produces better reconstructions.


\section{Experiments and Results}
\label{results}

\begin{figure}[htp]
\begin{center}
\subfigure
{\includegraphics[scale = 0.3]{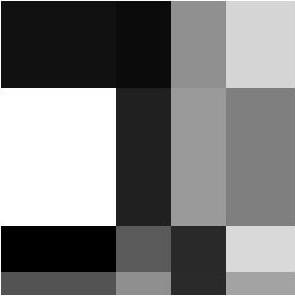}
\label{toy1}}
\subfigure
{\includegraphics[scale = 0.3]{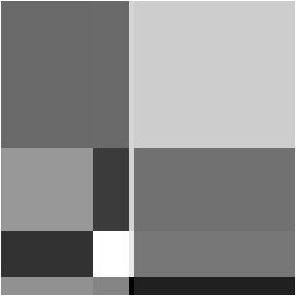}
\label{toy2}}
\caption{Sample of toy images used for testing}
\label{2dims}
\end{center}
\end{figure}

We evaluate the proposed approaches on 1-dimensional signals,  toy images (shown in Fig.~\ref{2dims}), and a real image. We used SpaRSA \cite{sparsa}, modified to suit the overlapping groups scenario and the associated modified case, to solve equations \eqref{oglr} and \eqref{moglasso}. We used the Haar wavelet basis as the sparsity inducing transform. Groups were defined according to the methods explained in section \ref{glasso}. 

\begin{figure}[!htp]
\begin{center}
\subfigure[lasso reconstruction (MSE=0.0043)]
{\includegraphics[scale = .45]{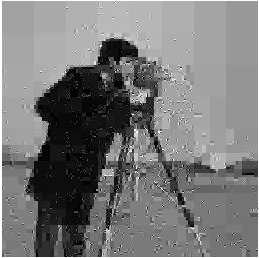}
\label{Lrec}}
\subfigure[OGLR reconstruction (MSE=0.0031)]
{\includegraphics[scale = .45]{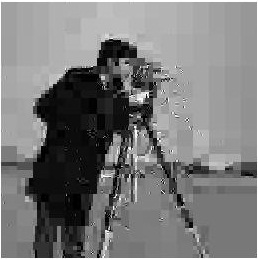}
\label{Glrec}}
\subfigure[lasso deblurring (MSE=0.010)]
{\includegraphics[scale = 0.45]{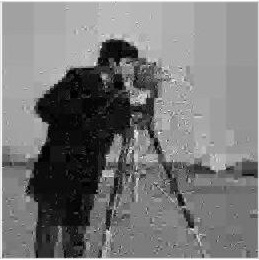}
\label{Ldec}}
\subfigure[OGLR deblurring (MSE=0.007)]
{\includegraphics[scale = 0.45]{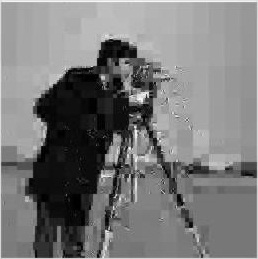}
\label{Gldec}}
\caption{Performance on the cameraman image}
\label{real}
\end{center}
\end{figure}

To illustrate the potential of the proposed methods we first consider compressed sensing and deconvolution results for the cameraman image, resized to $128 \times 128$.  
Fig. \ref{Lrec} and Fig. \ref{Glrec} show the compressed sensing results. The image was undersampled using a random \emph{iid} gaussian matrix, using only 800 samples for every $64 \times 64$ subimage. Fig. \ref{Ldec} and Fig. \ref{Gldec} show the results of deblurring the image, blurred with a gaussian kernel of variance 1.  The samples in both cases were corrupted by $WGN$ of variance 1.

\vspace{-5 mm}
\subsection{Varying the noise}
\begin{figure}[]
\begin{center}
\includegraphics[scale = 0.45]{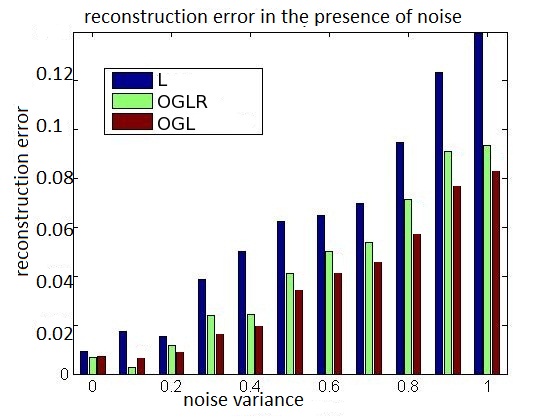}
\caption{Effect of varying the noise variance. Groups are formed according to Fig. \ref{PC4}}
\label{2dnoise}
\end{center}
\end{figure}

\begin{figure}[]
\begin{center}
\includegraphics[scale = 0.45]{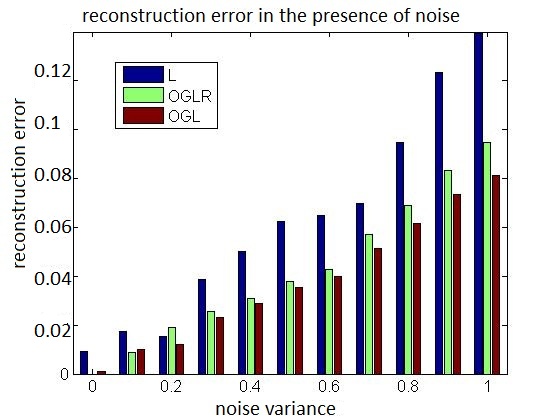}
\caption{Effect of varying the noise variance. Groups are formed according to Fig. \ref{PC2}}
\label{2dnoisepc}
\end{center}
\end{figure}

To evaluate the improvements of the group penalties we first consider performance relative to the signal-to-noise ratio. Our first set of experiments involved testing the recovery scheme in a compressed sensing framework (using an iid gaussian measurement matrix of size $800 \times 4096$). The noise variance was varied from 0 to 1 in steps of 0.1, and we measured the mean reconstruction error $||\theta^* - \hat{\theta}||_2^2$. Results obtained using the toy images are plotted in Fig.~\ref{2dnoise} and Fig.~\ref{2dnoisepc}.  The results are averaged over $100$ independent trials at each noise level.  A random `toy' image similar to those in Fig.~\ref{2dims} was generated for each trial.  We employed a grid search to pick the best value of $\lambda$ and $\tau$ (wherever applicable). Note that OGL and OGLR produces better results than standard lasso.

\vspace{-5 mm}
\subsection{Varying the number of measurements}
\begin{figure}[]
\begin{center}
\includegraphics[scale = 0.45]{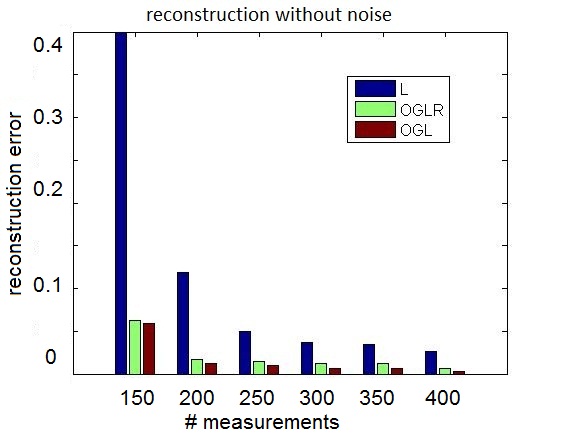}
\caption{Effect of varying the number of measurements taken. Note how the overlap lasso needs far fewer measurements than the lasso to achieve low errors. As the number of measurements is increased beyond 300, both methods stop improving significantly.}
\label{measurements}
\end{center}
\end{figure}

The group penalties also reduce the number of compressed sensing measurements needed to reconstruct images. To study this effect, we varied the number of rows of the matrix. The inputs used were random piecewise constant signals of length 1024, with at most 5 jumps assigned uniformly at random (the 1-D equivalent of Fig. \ref{2dims}), and the associated groups were determined using the binary tree structure of the 1-D DWT, with a group corresponding to a single parent-child pair (an edge of the tree as in Fig.~\ref{PC2}). It was observed that we need \emph{far} fewer measurements for robust recovery of signals when this implicit group structure is assumed, as opposed to that needed for the conventional lasso (see Fig. \ref{measurements}). 

\section{Conclusions}
\label{conc}

The proposed group penalties match the sparsity patterns of wavelet coefficients in natural images better than simple coordinate-wise penalties such as the $\ell_1$ (lasso)  penalty.  Like the $\ell_1$ penalty, linear inverse problems with group penalties are convex optimizations that can be solved efficiently and exactly.  Traditional Markov tree models for coefficients do not lead to convex optimizations. 
The non-separable nature of the group lasso (when groups are overlapping) was addressed by devising a new optimization criterion that can be solved by standard
methods based on separable penalties. The experiments demonstrate the performance gains of the group penalty methods compared to the lasso.

\section{Acknowledgements}
\label{ack}
This work was partially supported by the DARPA / AFRL A2I program.

\bibliographystyle{plain}
\bibliography{icip_RDN_Jan20}

\end{document}